\documentclass[runningheads]{llncs}
\usepackage{graphicx}

\usepackage{tikz}
\usepackage{comment}
\usepackage{amsmath,amssymb} 
\usepackage{color}
\usepackage{subcaption}
\usepackage{makecell}
\usepackage[accsupp]{axessibility}  

\usepackage{booktabs}
\usepackage{floatrow}
\newfloatcommand{capbtabbox}{table}[][\FBwidth]
\usepackage{blindtext}
\usepackage[T1]{fontenc}
\usepackage[font=small,labelfont=bf,tableposition=top]{caption}
\DeclareCaptionLabelFormat{andtable}{#1~#2  \&  \tablename~\thetable}
\usepackage{multirow}

\begin{document}
\pagestyle{headings}
\mainmatter

\title{Non-Autoregressive Sign Language Production \\via Knowledge Distillation} 

\titlerunning{Abbreviated paper title}
%
\author{Eui Jun Hwang \and Jung Ho Kim \and Suk Min Cho \and Jong C. Park}
\authorrunning{Hwang et al.}
%
\institute{Korea Advanced Institute of Science\\
    and Technology (KAIST)\\
    Daejeon, Korea
\email{\{ejhwang,jhkim,park\}@nlp.kaist.ac.kr}}
\maketitle

\begin{abstract}
Sign Language Production (SLP) aims to translate expressions in spoken language into corresponding ones in sign language, such as skeleton-based sign poses or videos. Existing SLP models are either AutoRegressive (AR) or Non-Autoregressive (NAR). However, AR-SLP models suffer from regression to the mean and error propagation during decoding. NSLP-G, a NAR-based model, resolves these issues to some extent but engenders other problems. For example, it does not consider target sign lengths and suffers from false decoding initiation. We propose a novel NAR-SLP model via Knowledge Distillation (KD) to address these problems. First, we devise a length regulator to predict the end of the generated sign pose sequence. We then adopt KD, which distills spatial-linguistic features from a pre-trained pose encoder to alleviate false decoding initiation. Extensive experiments show that the proposed approach significantly outperforms existing SLP models in both Fr\'{e}chet Gesture Distance and Back-Translation evaluation.

\end{abstract}

\section{Introduction} \label{sec:intro}
Sign language (SL), a rich multi-channel language, is the primary form of communication in the Deaf communities. However, unlike spoken languages, manual and non-manual components such as hands, face, and body play essential roles in making effective communication~\cite{sandler2006sign}. This aspect of SL gives rise to a communication gap between SL users and spoken language users. Therefore, many researchers have proposed various methods to convert expressions in spoken language into corresponding ones in sign language to fill this gap.

\begin{figure}
    \centering
    \includegraphics[width=0.9\columnwidth]{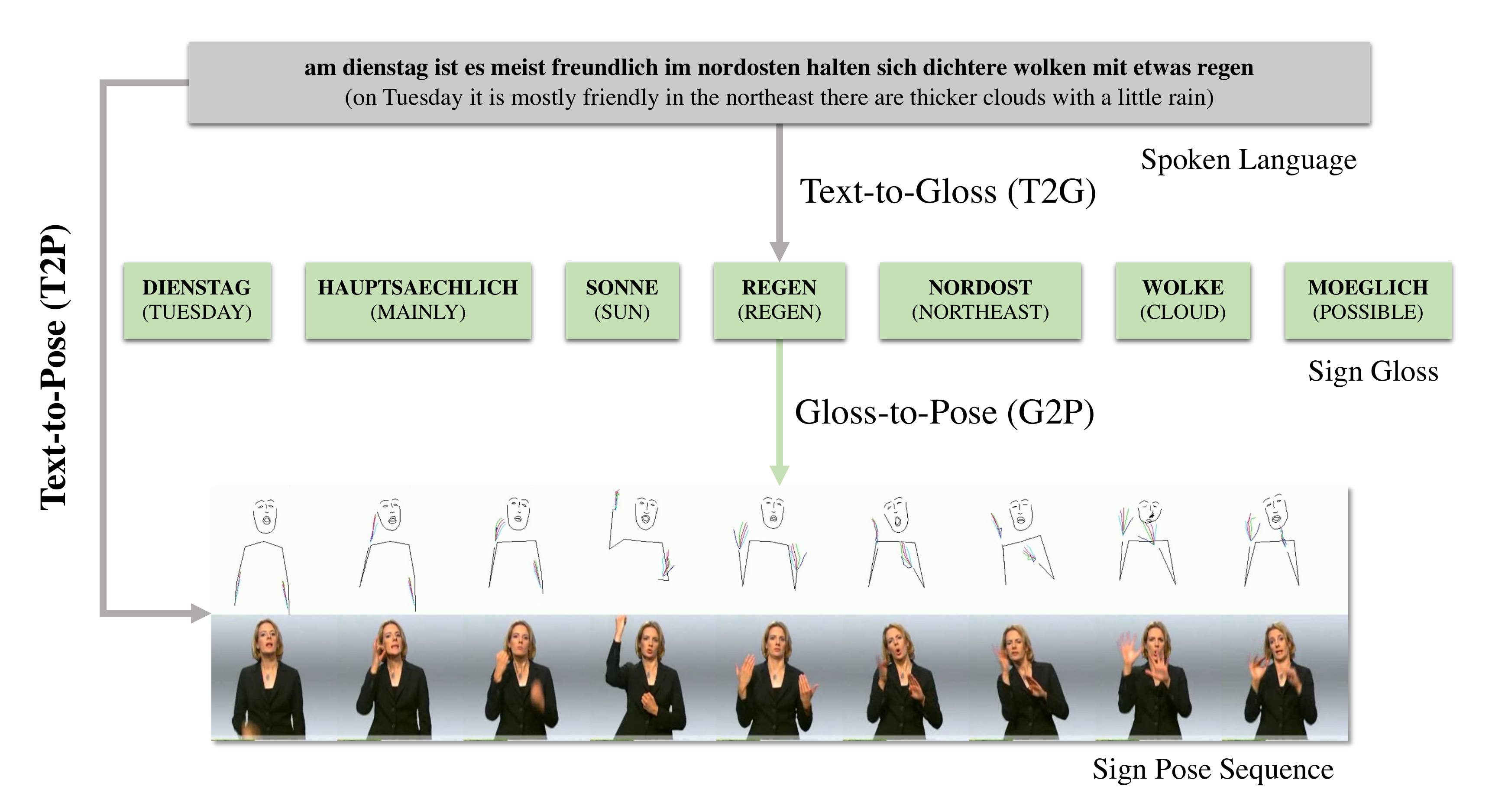}
    \caption{An overview of Sign Language Production (SLP). There are two different tasks: Gloss-to-Pose (G2P) and Text-to-Pose (T2P). Recent SLP works demonstrate that T2P performs better than G2P. Hence, we focus on the T2P task in this work.}
    \label{fig:1}
\end{figure}

Sign Language Production (SLP) aims to translate expressions in spoken language into corresponding ones in SL, such as sign skeleton-based~\cite{saunders2020progressive,saunders2021mixed,Hwang2021nonautoregressive}, avatar-based~\cite{kacorri2016selecting} sign poses, and videos~\cite{stoll2018sign}. As shown in Figure~\ref{fig:1}, recent SLP works have been carried out either gloss-to-pose (G2P) or text-to-pose (T2P). G2P does not use text but only glosses as input to produce the sign pose sequence. On the other hand, T2P produces the sign pose sequence directly from the given text without intermediate glosses\footnote{Glosses are a simplified notation that connects spoken language and SL.}. According to recent SLP studies~\cite{saunders2021mixed,Hwang2021nonautoregressive}, T2P has achieved better performance than G2P, suggesting that gloss representation may be losing some rich information in sign language expressions. Therefore, in this work, we focus on T2P to overcome the limitations of the gloss representation.

Most existing approaches to T2P use an AutoRegressive (AR) model based on direct mapping to generate the current pose depending on the previous one. However, these approaches often regress to the mean, propagate errors during decoding and suffer from the slow decoding procedure~\cite{huang2021towards}. To address these limitations, Gaussian Mixture Network~\cite{saunders2021continuous} and Mixture of Motion Primitives~\cite{saunders2021mixed} have been proposed. While these efforts have mitigated the limitations to some extent, the fundamental problem posed by AR approaches has not yet been fully resolved.

To solve the problems in AR models, Huang et al.~\cite{huang2021towards} and Hwang et al.~\cite{Hwang2021nonautoregressive} proposed NAT-EA and NSLP-G, respectively, which work in a NAR manner. NAT-EA is designed for the G2P task and consists of a transformer-based encoder, spatial-temporal graph pose generator, and a length regulator. NSPL-G has achieved the goal with two learning steps: learning the spatial-linguistic aspect of SL using Variational Autoencoder (VAE)~\cite{kingma2013auto} and mapping text into the spatial linguistic feature using NAR Transformer. Note that NAT-EA is designed for the G2P task only and is not considered in this paper.

Although NSLP-G~\cite{Hwang2021nonautoregressive} outperforms the previous AR models, it still has several limitations. (1) The model does not include a component to predict the target length but instead introduces a masked loss as a partial solution. (2) It performs worse in a short range of prediction. We call it ``false decoding initiation''. (3) VAE has been optimized with Mean Squared Error (MSE) loss. However, MSE is not relative and tends to lose its signal when processing perceptually important signals~\cite{wang2009mean}, making it difficult to produce detailed outputs such as hand representations in SL. 

\begin{figure}[t!]
    \centering
    \includegraphics[width=0.9\textwidth]{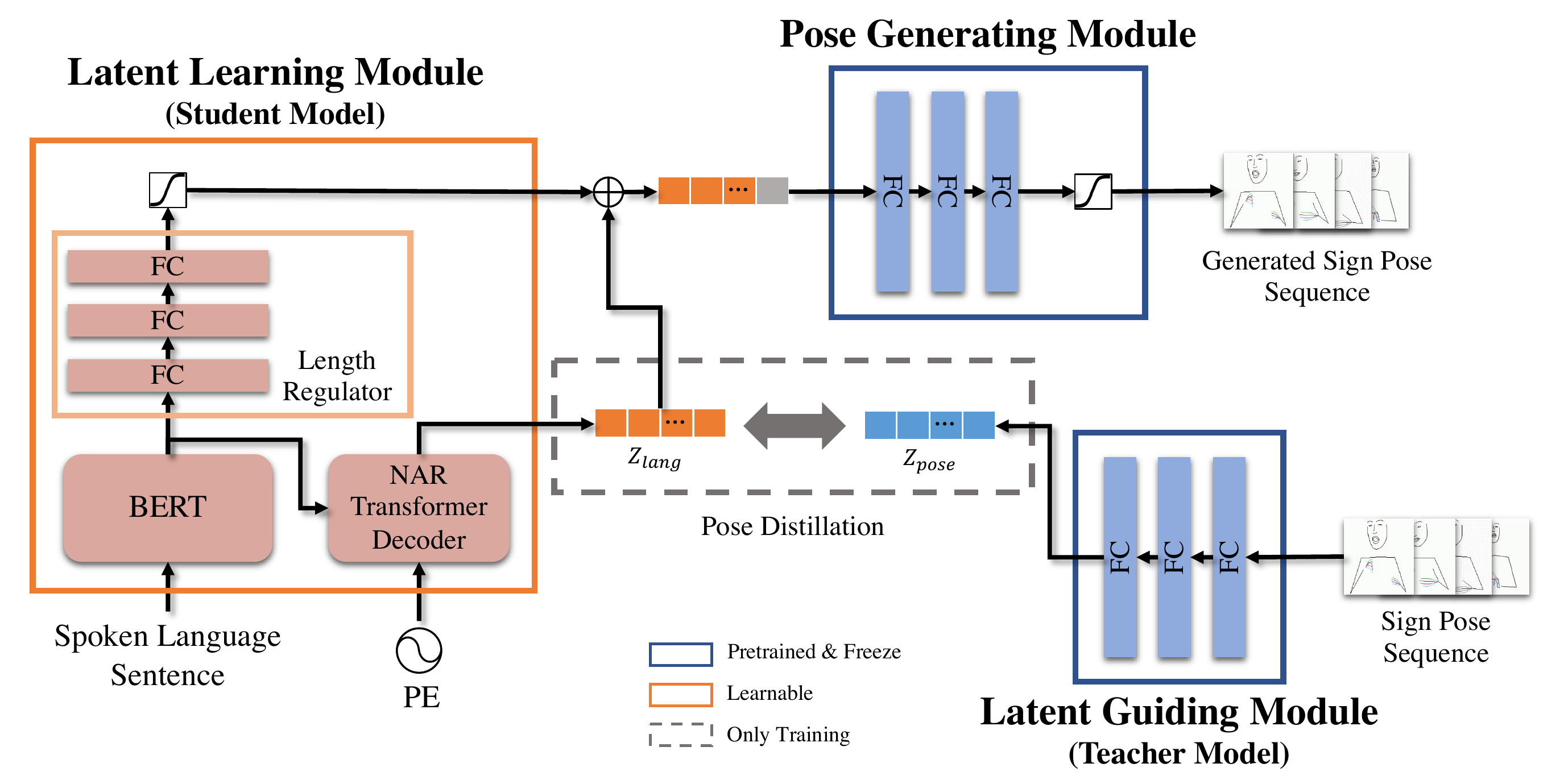}
    \caption{An overview of the proposed NAR-SLP model. To apply Knowledge Distillation (KD), we pre-trained a sign pose network based on VAE. The trained encoder and decoder are used as Latent Guiding Module (LGM) and Pose Generating Module (PGM), respectively. Latent Learning Module (LLM) generates a length-regulated sequence of latent representation. $\oplus$ denotes a length regulating operation. LLM consists of three sub-modules: BERT as a language embedding module, Transformer-based NAR decoder, and Length Regulator. During training, as a pose distillation process, LGM acts as a teacher network, teaching LLM to generate $z_{lang}$ similar to $z_{pose}$.}
    \label{fig:2}
\end{figure}

We propose a novel NAR-SLP model via Knowledge Distillation (KD)~\cite{hinton2015distilling} as shown in Figure~\ref{fig:2}. To resolve (1), our model is designed to generate length-regulated sign poses by introducing a length regulator. A pre-trained language model such as BERT~\cite{Devlin2018} is employed for increased modality in our model. To resolve (2), we introduce KD, which aligns language features directly with sign pose features. This alleviates the false decoding initiation by a large margin throughout experiments. To resolve (3), we assume that the final output of the model follows a multivariate Bernoulli distribution. The sign poses can therefore be interpreted as an estimate of the probability that a particular point is attended by an observer. Hence, we optimize VAE using Binary Cross Entropy (BCE) loss~\cite{de2005tutorial}.

The difference from the previous NAR model is the exact formulation of the length regulator and application of KD. In particular, KD addresses the performance degradation due to increased modality in our model as well as false decoding initiation. Our main contributions are summarized as follows: 
\begin{itemize}
    \item[$\circ$] We introduce a novel Non-Autoregressive Sign Language Production (NAR-SLP) model via Knowledge Distillation (KD) for the T2P task. This is the first SLP model to apply the KD approach.
    \item[$\circ$] We design the proposed model to generate length-regulated outputs to fill the gap in the previous NAR-SLP model.
    \item[$\circ$] Our approach improves performance by addressing the limitations of previous SLP models.
    \item[$\circ$] Extensive experiments demonstrate the effectiveness of the proposed model on the PHOENIX-2014T dataset.
\end{itemize}

\section{Related Work}

\subsubsection{Non-Autoregressive Models.}
AR-based models have achieved great success in machine translation~\cite{vaswani2017attention}. Typically, they generate outputs sequentially where the current output is dependent on the previous one. In the case of predicting the human pose, however, AR models are prone to converge to a mean pose, which hinders predictions of realistic poses~\cite{li2018convolutional}. Several efforts have been made to avoid AR modeling. Gu et al.~\cite{gu2017non} propose Non-Autoregressive Transformer (NAT) that uses a fertility predictor to represent the number of times each source token is copied to predict the target length. Li et al.~\cite{li2021multitask} use the NAR decoder to independently generate each human pose given context features from the encoder and positional information.

\subsubsection{Knowledge Distillation.}
KD is a method of distilling knowledge from a teacher model to a student for model compression and acceleration~\cite{hinton2015distilling}, a method frequently used in various fields, such as natural language processing and visual recognition~\cite{gou2021knowledge}. Wang et al.~\cite{Wang_2021_CVPR} exploit KD for guiding the pose estimator to learn clean pose knowledge during training on hard samples with noise. Also, Khan et al.~\cite{raj-khan-etal-2021-towards-developing} propose a KD framework for transferring the knowledge from a pre-trained linguistic vision models to multilingual BERT.

\subsubsection{Sign Language Production.}
Previous SLP works have started with avatar-based approaches. They can produce human-like signs but rely on phrase look-ups and predefined motion dictionaries~\cite{mcdonald2016automated}, or require expensive motion capture or pre-recorded phrases~\cite{lu2011data}. With recent advances in deep learning, Stoll et al.~\cite{stoll2018sign} propose the first SLP model to translate text into glosses and map them to corresponding sign poses. Zelinka and Kanis~\cite{zelinka2020neural} propose the first end-to-end SLP model with a fixed length. They also propose a gradient descent method for skeletal refinement. Recently, Jiang et al.~\cite{jiang2021skeletor} propose a transformer-based skeletal refinement model that refines the 3D lifted skeleton with a BERT-like training approach. Saunders et al.~\cite{saunders2020progressive} propose Progressive Transformer (PT), which uses a counter-encoding scheme to directly learn the mapping between spoken language and sign pose sequence. As follow-up studies, a Mixed Density network~\cite{saunders2021continuous} and a Mixture of Experts~\cite{saunders2021mixed} are applied to PT. For the G2P task, Huang et al.~\cite{huang2021towards} propose a NAR-SLP model with an External Aligner for sequence alignment learning. Hwang et al.~\cite{Hwang2021nonautoregressive} propose a NAR-SLP model for both G2P and T2P, which adopts indirect mapping using pre-trained VAE.

\section{Approach} \label{sec:method}
Our proposed approach is shown in Figure~\ref{fig:2}. We utilize a VAE model trained in individual sign poses, similar to NSLP-G. However, while NSLP-G only uses a pre-trained decoder to align the outputs of a NAR Transformer, both encoder and decoder are used in our model. The trained encoder is used as a teacher network called Latent Guiding Module (LGM), and the trained decoder is used as Pose Generating Module (PGM). Latent Learning Module (LLM), a student network, generates a length-regulated sequence of latent representation from the given spoken language sentence. LLM learns how to generate an appropriate latent representation for sign pose generation in the teacher network across the different modalities between spoken language and sign poses.

In the following subsections, we will first define the problem of SLP and then cover each module in detail, followed by a training scheme.

\subsection{Problem Definition}
Given a sequence of spoken words $X=(x_1, x_2, ..., x_u)$, the goal of SLP is to generate a consecutive sign pose sequence with length $T$, $Y=(y_1, y_2, ..., y_T)\in \mathbb{R}^{T\times J\times K}$, where each frame $y_t=\{y_{t}^{j}\}_{j=1}^{J}$ represents a single sign pose, containing $J$ joint data. $y_{t}^{j}\in \mathbb{R}^K$ is a minimal per-joint representation at the $t$-th frame and the $j$-th joint, of which $K$ is the feature dimension which presents the sign pose joint data.

Our goal is to train a NAR Transformer to generate the realistic sign pose sequence $Y$ with a corresponding length $T$ from the given words $X$. We introduce two different conditional probabilities, $P(Z|X)$ and $g(Y|Z)$, as in the previous work~\cite{Hwang2021nonautoregressive}. However, their work does not truly satisfy $P(Z|X)$ because there is no part for the target length $T$. Therefore, we introduce revised $P(Z|X)$ to achieve our goal. More details are covered in Section~\ref{sec:lln}. We stipulate that the model has $Z=(z_1,z_2,...,z_T)$ as an intermediate representation between spoken words $X$ and sign pose sequence $Y$.

\subsection{Latent Guiding and Pose Generating Modules}
We first train VAE on individual sign poses. It can be represented as:
\begin{equation}
    enc(y)=q(z|y),\;dec(z)=p(y|z),
\end{equation}
where $y$ denotes a single sign pose skeleton and $z$ denotes an encoded sign pose latent.

More specifically, $y$ is encoded as sign pose distribution parameters $\mu$ and $\sigma$ of the sign pose distribution. We use reparameterization~\cite{kingma2013auto} to sample a latent space $z\in {\mathbb{R}^{d}}$ from this distribution. The decoder is used to reconstruct the original sign pose $y$, given the corresponding $z$. The objective of VAE is formulated as:
\begin{equation}
    \mathcal{L}_{vae}(y)=-\mathbb{E}_{q(z|y)}[log{p}(y|z)] +\lambda KL(q(z|y)||p(z)),
\end{equation}
where $p(z)$ is the prior distribution, $KL$ is the Kullback-Leibler divergence, and $\lambda$ is the hyper-parameter to control the balance of losses. The first term allows the model to encode the sign pose $y$ into the latent space $z$ for reconstruction. The second term pushes posterior distribution to be close to the prior distribution. After the learning process, the trained encoder and decoder are used as the teacher network LGM and PGM, respectively.

\subsection{Latent Learning Module} \label{sec:lln}
We propose LLM to model $P(Z|X)$. As shown in Figure~\ref{fig:2}, it has different sub-modules: a spoken language encoder, length regulator, and NAR decoder. The previous work~\cite{Hwang2021nonautoregressive} achieves the SLP model non-autoregressively but does not include a conditional probability of the target length. Instead, they introduce a masked loss to solve this problem partially. To fill this gap, we redefine $P(Z|X)$ as follows:
\begin{equation}\label{eq:eq3}
    p_{\mathcal{NA}}(Z|X)=p_{L}(T|x_{1:U})\cdot \prod_{t=1}^{T}p(z_t|x_{1:U}),
\end{equation}
where $p_{L}(T|x_{1:U})$ is a conditional probability of a target length $T$. More details of each sub-module are described in the following paragraphs.

\subsubsection{Language Encoder and Length Regulator.}
We employ BERT~\cite{Devlin2018} as the spoken language encoder. BERT is trained on a very large unlabeled corpus and is successfully applied to the SLT task~\cite{miyazaki2020machine}. 

To satisfy the first term $p_{L}(T|x_{1:U})$ in Equation~\ref{eq:eq3}, we employ a simple network consisting of multiple fully-connected (FC) layers with a Sigmoid activation at the end of the layer to adjust the length of the generated sequence of latent representation. Specifically, we design the length regulator to generate the ratio of target sign length to maximum target length using [CLS] token, which is widely used in BERT-based classification tasks~\cite{yu2019improving,gao2019target}.

\subsubsection{Non-AutoRegressive Decoder.}
Our model uses a NAR Transformer decoder but removes the AR connection. It is based solely on an attention mechanism to generate representations of entire sequences with global dependencies. It is formulated as:
\begin{equation} \label{eq:attn}
    Attention(Q,K,V)=softmax(\frac{QK^T}{\sqrt{d_k}})V,
\end{equation}
where $Q$, $K$ and $V$ are query, key and value, respectively. Finally, Multi-Head Attention (MHA) can be formulated as:
\begin{align}\label{eq:mha}
    & MHA(Q,K,V) = Concat({head}_i, ..., {head}_n)W_O,\\
    & head_i = Attention(QW^i_Q,KW^i_K,VW^i_V),
\end{align}
where $W_Q$,$W_K$, and $W_V$ are weights related to each input.

Given the embedded input and temporal information, the NAR decoder generates a sequence of latent $Z_{lang}$ of the given spoken words $X$. Specifically, it takes Positional Encoding (PE) as $Q$ and the encoder output as $K$ and $V$, respectively.

\subsection{Training}
We define several loss terms to train our model in this section and present an ablation study in Section~\ref{sec:ablation}.

\subsubsection{Pose Loss.}
Previous work~\cite{Hwang2021nonautoregressive} uses MSE loss to train the VAE. However, MSE tends to lose its signal when processing perceptually important signals~\cite{wang2009mean}. As a result, the model does not effectively capture detailed outputs such as hand movement and facial expressions, which are relatively a small variance compared to body movements within the dataset.

To address this problem, we normalize the ground-truth joints so that each value is in the range $[0,1]$. The sign poses can therefore be interpreted as an estimate of the probability that a particular joint is attended by an observer. As with the work presented in~\cite{kingma2013auto}, it is tempting to induce a multivariate Bernoulli distribution. Thus, Sigmoid has been applied element-wise and BCE loss is used to measure the difference between the ground-truth sign pose sequence $Y$ and generated sign pose sequence $\hat{Y}$, defined as: 
\begin{equation}
    \mathcal{L}_{pose}=-\frac{1}{N}\sum_{i=1}^{N}{(y_ilog(\hat{y}_i)}+(1-y_i)log(1-\hat{y}_i)),
\end{equation}
where $\hat{y}_i$ denotes the predicted joint and $y_i$ denotes the corresponding target joint from the ground-truth.

On the other hand, MSE loss is used to optimize LLM through the pre-trained pose generator PGM indirectly. This is because we assume that the NAR decoder does not follow the Bernoulli distribution. Hence, it can be defined as:
\begin{equation}
    \mathcal{L}_{pose}=\frac{1}{N}\sum_{i=1}^{N}(y_i-\hat{y}_i)^2.
\end{equation}

\subsubsection{Distillation Loss.}
We argue that input language features should be directly aligned with the sign pose features to generate a sign pose sequence correctly. This helps the model to represent the cross-modality between language and sign poses. We compute a pose distillation loss, which directly measures the difference between $Z_{lang}$ and $Z_{pose}$. It plays a significant role in our model throughout experiments and can be defined as:
\begin{equation}
    \mathcal{L}_{distil}=\frac{1}{N}\sum_{i=1}^{N}(z_{i}^{pose}-z_{i}^{lang})^2.
\end{equation}

\subsubsection{Length Loss.}
As described in Section~\ref{sec:lln}, our model is designed to generate length-regulated outputs. Specifically, the regulator determines how many frames should be discarded from the generated latent sequence. Therefore, BCE loss is used to measure the difference between the ground-truth length ratio $L$, and the predicted length ratio $\hat{L}$, defined as:
\begin{equation}
    \mathcal{L}_{length}=-\frac{1}{N}\sum_{i=1}^{N}{(l_ilog(\hat{l}_i)}+(1-l_i)log(1-\hat{l}_i)).
\end{equation}

\subsubsection{}
The resulting total loss is defined as the summation of different terms, as:
\begin{equation}\label{eq:loss}
    \mathcal{L}=\mathcal{L}_{pose}+\mathcal{L}_{distil}+\mathcal{L}_{length},
\end{equation}
where the loss terms are assigned the same weight empirically.

\section{Experimental Settings} \label{sec:expenv}
In this section, we describe the dataset and prepossessing, followed by the implementation details of our model. We then introduce baselines in our experiments.

\subsection{Dataset and Preprocessing}
We use the publicly available RWTH-PHOENIX-Weather 2014T dataset~\cite{camgoz2018neural}. To the best of our knowledge, the current trend in SLP works is using one method on a single dataset due to the scarcity of SLP datasets. The PHOENIX-2014T dataset was chosen because it has been extensively and actively used in sign language research. This dataset contains 8,257 pairs of German and German Sign Language (DGS) videos with word-level annotations, collected from weather forecast of PHOENIX TV station. It includes 2,887 different German words and 1,066 different DGS glosses. We use OpenPose~\cite{cao2019openpose} to extract manual and non-manual features. The manual features are lifted into 3D using the skeletal correction model~\cite{zelinka2020neural}. Recently, Duarte et al.~\cite{duarte2021how2sign} introduced a new benchmark dataset for SL field. However, since the dataset is not yet fully released, we do not include results on the dataset.

\subsection{Implementation Details}
For all our experiments, LGM and PGM have 3 linear layers with ReLU and PGM has Sigmoid at the end of the layer, with $\lambda$ set to $0.0001$. In LLM, we set the embedding dimension to 768, the number of layers to 8, the number of heads in multi-head attention to 8, the dropout rate to 0.1, and the dimension of the intermediate feedforward network to 2,048. All parts of our network are trained with Xavier initialization and AdamW optimization, with a learning rate of $0.0002$. Our model is implemented using PyTorch Lightning~\cite{falcon}. Training takes 24 hours for 500 epochs on a single Tesla V100 GPU, using 40GB GPU memory with batch size 40.

\subsection{Baselines}
We compare our approach to several other methods, including AR and NAR state-of-the-art models and other baseline models. Note that these models are trained for 500 epochs for a fair comparison. Overall, more epochs improve performance, but we stop training to keep computational costs low.

\subsubsection{Ground Truth (GT).}
We use sign poses of human signers in the PHOENIX-2014T dataset. Note that we considered the sign poses extracted from Openpose as the ground truth.

\subsubsection{Mean and Random.}
To set the bottom baselines, we employ two different models, Mean and Random, each generating the mean pose and a randomly selected pose sequence from the training set.

\subsubsection{Progressive Transformer (PT).}
Since PT~\cite{saunders2020progressive} is the only publicly available AR-SLP model, we compare the T2P model of PT with our approach. Note that the results they presented are not comparable to ours as the authors did not release a pre-trained BT model and a full sign pose joint data, including facial landmarks. Therefore, we reproduce the results of the T2P models using base, and Future Prediction and Gaussian noise (FP\&GN) settings, which shows the best performance in their paper.

\subsubsection{NSLP-G.}
NSLP-G~\cite{Hwang2021nonautoregressive} is the first NAR-based model to apply an unsupervised learning approach. We compare our approach with their T2P models using base, gloss supervision (GS), and fine-tuning (ft) settings. The results presented in their work are not the same as ours. This is because, for FGD, we use BCE loss to optimize the evaluation model for the reasons stated in Section~\ref{sec:method}. 

\section{Experimental Results}
We conduct extensive experiments to evaluate our approach and compare it against the aforementioned baselines. The experiments are designed to validate the proposed solutions to the limitations described in Section~\ref{sec:intro}. In this section, we describe the evaluation metrics, followed by quantitative and qualitative results.

\subsection{Evaluation metrics}

\subsubsection{Fr\'{e}chet Gesture Distance.}
We evaluate the visual quality and realism of the generated sign pose sequence using Fr\'{e}chet Gesture Distance (FGD)~\cite{yoon2020speech}. It is based on the concept of Fr\'{e}chet Inception Distance (FID)~\cite{heusel2017gans}, and measures how close the distribution of generated sign pose sequence $\hat{Y}$ is to real sign pose sequence $Y$. It can be formulated as:
\begin{equation}
    FGD(Y, \hat{Y}) = {\left \| \mu_r - \mu_g \right \|}^{2} + Tr(\Sigma_r + \Sigma_g - 2(\Sigma_r\Sigma_g)^{\frac{1}{2}}),
\end{equation}
where $\mu_{r}$ and $\Sigma_{r}$ are the first and second moments of the latent feature distribution $Z_{r}$ of real sign poses $Y$, respectively, and $\mu_{g}$ and $\Sigma_{g}$ are the first and second moments of the latent feature distribution $Z_{g}$ of generated sign poses $\hat{Y}$, respectively. Transformer-based Autoencoder (TAE)~\cite{Hwang2021nonautoregressive} is used as an evaluation model for FGD. It is trained on the PHOENIX-2014T dataset with a fixed length of sign pose sequences. We set it to 34. 

\subsubsection{Back-Translation.}
We apply the Back-Translation (BT) method proposed in~\cite{saunders2020progressive} as a means of BLEU and ROUGE evaluation. BT uses pre-trained Sign Language Translation (SLT) model to translate the generated sign pose sequence into spoken language or sign glosses. We use a state-of-the-art SLT model~\cite{camgoz2020sign} modified to take a sign pose sequence as an input. It is trained on the PHOENIX-2014T dataset.

\begin{table}[t!]
\centering
\setlength{\tabcolsep}{3pt}
\resizebox{\columnwidth}{!}{%
\begin{tabular}{lcccccccccccc}
\toprule
\multicolumn{1}{l|}{\multirow{2}{*}{Models}} &
  \multicolumn{6}{c|}{DEV} &
  \multicolumn{6}{c}{TEST} \\
\multicolumn{1}{c|}{} &
  FGD$\dagger$ &
  BLEU-1 &
  BLEU-2 &
  BLEU-3 &
  BLEU-4 &
  \multicolumn{1}{c|}{ROUGE} &
  FGD$\dagger$ &
  BLEU-1 &
  BLEU-2 &
  BLEU-3 &
  BLEU-4 &
  ROUGE \\ \midrule
\multicolumn{1}{l|}{GT} &
  0.0 &
  32.28 &
  19.51 &
  13.93 &
  10.96 &
  \multicolumn{1}{c|}{32.47} &
  0.0 &
  32.48 &
  19.57 &
  13.83 &
  10.71 &
  32.22 \\ \midrule
\multicolumn{1}{l|}{Mean} &
  0.49 &
  13.93 &
  3.07 &
  0.98 &
  0.40 &
  \multicolumn{1}{c|}{14.28} &
  0.56 &
  13.25 &
  2.84 &
  0.84 &
  0.29 &
  14.40 \\
\multicolumn{1}{l|}{Random} &
  0.44 &
  17.36 &
  5.04 &
  1.94 &
  0.91 &
  \multicolumn{1}{c|}{16.40} &
  0.48 &
  18.13 &
  5.75 &
  2.20 &
  0.92 &
  17.46 \\ \midrule
\multicolumn{13}{c}{\textit{Autoregressive   Model}} \\ \midrule
\multicolumn{1}{l|}{\textit{PT}} &
  \textit{} &
  \textit{} &
  \textit{} &
  \textit{} &
  \textit{} &
  \multicolumn{1}{c|}{\textit{}} &
  \textit{} &
  \textit{} &
  \textit{} &
  \textit{} &
  \textit{} &
  \textit{} \\
\multicolumn{1}{l|}{\hspace*{0.3cm}Base} &
  1.01 &
  18.47 &
  5.07 &
  1.90 &
  0.97 &
  \multicolumn{1}{c|}{17.26} &
  0.97 &
  17.55 &
  4.59 &
  1.62 &
  0.73 &
  17.38 \\
\multicolumn{1}{l|}{\hspace*{0.3cm}FP\&GN} &
  0.33 &
  17.28 &
  5.49 &
  2.65 &
  1.63 &
  \multicolumn{1}{c|}{16.26} &
  0.32 &
  16.52 &
  4.24 &
  1.40 &
  0.58 &
  16.30 \\ \midrule
\multicolumn{13}{c}{\textit{Non-Autoregressive Model}} \\ \midrule
\multicolumn{1}{l|}{\textit{NSLP-G*}} &
  \textit{} &
  \textit{} &
  \textit{} &
  \textit{} &
  \textit{} &
  \multicolumn{1}{c|}{\textit{}} &
  \textit{} &
  \textit{} &
  \textit{} &
  \textit{} &
  \textit{} &
  \textit{} \\
\multicolumn{1}{l|}{\hspace*{0.3cm}Base} &
  0.24 &
  28.24 &
  15.01 &
  9.85 &
  7.35 &
  \multicolumn{1}{c|}{27.19} &
  0.26 &
  28.60 &
  15.89 &
  10.78 &
  8.23 &
  27.94 \\
\multicolumn{1}{l|}{\hspace*{0.3cm}GS} &
  0.25 &
  25.97 &
  13.19 &
  8.62 &
  6.45 &
  \multicolumn{1}{c|}{25.32} &
  0.28 &
  26.92 &
  14.04 &
  9.30 &
  7.03 &
  26.92 \\
\multicolumn{1}{l|}{\hspace*{0.3cm}Base (+ft)} &
  \underline{0.22} &
  \underline{33.01} &
  \underline{19.67} &
  \underline{14.02} &
  \underline{10.97} &
  \multicolumn{1}{c|}{\underline{32.21}} &
  \underline{0.23} &
  \underline{32.63} &
  \underline{19.42} &
  \underline{13.47} &
  \underline{10.39} &
  \underline{32.24} \\
\multicolumn{1}{l|}{\hspace*{0.3cm}GS (+ft)} &
  0.25 &
  30.62 &
  16.75 &
  10.92 &
  8.13 &
  \multicolumn{1}{c|}{29.41} &
  0.27 &
  29.89 &
  16.57 &
  11.03 &
  8.26 &
  29.16 \\
\multicolumn{1}{l|}{\textit{Ours}} &
  \multicolumn{1}{l}{} &
  \multicolumn{1}{l}{} &
  \multicolumn{1}{l}{} &
  \multicolumn{1}{l}{} &
  \multicolumn{1}{l}{} &
  \multicolumn{1}{l|}{} &
  \multicolumn{1}{l}{} &
  \multicolumn{1}{l}{} &
  \multicolumn{1}{l}{} &
  \multicolumn{1}{l}{} &
  \multicolumn{1}{l}{} &
  \multicolumn{1}{l}{} \\
\multicolumn{1}{l|}{\hspace*{0.3cm}w/o KD} &
  0.32 &
  25.54 &
  12.82 &
  8.70 &
  6.62 &
  \multicolumn{1}{c|}{25.87} &
  0.36 &
  25.91 &
  13.56 &
  9.14 &
  6.90 &
  26.24 \\
\multicolumn{1}{l|}{\hspace*{0.3cm}w/ KD} &
  \textbf{0.15} &
  \textbf{36.81} &
  \textbf{23.90} &
  \textbf{17.50} &
  \textbf{13.90} &
  \multicolumn{1}{c|}{\textbf{36.89}} &
  \textbf{0.16} &
  \textbf{37.43} &
  \textbf{24.50} &
  \textbf{17.83} &
  \textbf{13.92} &
  \textbf{37.26} \\ \bottomrule
\end{tabular}%
}
\caption{We compare the recent SLP work on the PHOENIX-2014T dataset using FGD (lower is better), BLEU and ROUGE metrics. Note that, due to differences in implementation (lifted sign pose data, number of epochs used), the metrics for the baselines differ from those reported in their paper. * indicates that we test the model under a fair setting, and $\dagger$ indicates scaled $10\times$ for better readability. ft denotes finetuning of the NSLP-G model, and KD denotes Knowledge Distillation of our model.}
\label{tab:tab1}
\end{table}

\begin{table}[t!]
\centering
\setlength{\tabcolsep}{8pt}
\resizebox{1.0\columnwidth}{!}{%
    \begin{tabular}{ccccc|cccc}
    \toprule
    \multicolumn{1}{l|}{\multirow{3}{*}{Models}} & 
    \multicolumn{4}{c|}{Intervals (Seconds*)} & \multicolumn{4}{c}{Short Intervals (Seconds*)} \\ 
    \multicolumn{1}{c|}{}                        & \makecell{0$\sim$50 \\ (0$\sim$2s)} & \makecell{50$\sim$100 \\ (2$\sim$4s)} & \makecell{100$\sim$150 \\ (4$\sim$6s)} & \makecell{150$\sim$200 \\ (6$\sim$8s)} & \makecell{$\sim$20 \\ ($\sim$0.8s)} & \makecell{$\sim$40 \\ ($\sim$1.6s)} & \makecell{$\sim$60 \\ ($\sim$2.4s)} & \makecell{$\sim$80 \\ ($\sim$3.2s)} \\ \midrule
    \multicolumn{9}{c}{\textit{Autoregressive}}                                                                       \\ \midrule
    \multicolumn{1}{l|}{PT}                      & 0.35       & 0.91         & 1.17          & 1.18         & \textbf{0.04}         & \textbf{0.23}   & 0.46                  & 0.60 \\
    \multicolumn{1}{l|}{PT(FN\&GN)}              & \underline{0.24}       & 0.29         & 0.36          & 0.37          & \underline{0.05}                  & 0.25            & \underline{0.23}                  & \underline{0.23} \\ \midrule
    \multicolumn{9}{c}{\textit{Non-Autoregressive}}                                                                   \\ \midrule
    \multicolumn{1}{l|}{NSLP-G (base+ft)}        & 0.45       & \underline{0.28}         & \underline{0.16}          & \underline{0.11}          & 0.50                  & 0.60            & 0.42                  & 0.39  \\
    \multicolumn{1}{l|}{Ours (w/ KD)} & \textbf{0.21} & \textbf{0.18} & \textbf{0.12} & \textbf{0.09}  & 0.09                  & \underline{0.24}            & \textbf{0.22}         & \textbf{0.20} \\ \bottomrule
    \end{tabular}
}

\caption{FGD results of comparison with state-of-the-art models according to each frame length interval. * indicates that the generated frames are assumed to be 25 FPS. Lower numbers indicate better performance. Bold is the best, and underlined is second.}
\label{tab:2}
\end{table}

\subsection{Quantitative Results}

\subsubsection{Overall Assessment.}
As shown in Table~\ref{tab:tab1}, our method shows the best performance on all criteria. FGD of ours is 0.15, which is significantly lower than other methods. This indicates that the model can generate outputs much closer to a real sign pose sequence. It also achieves 13.90 and 36.89 in BLEU-4 and ROUGE, respectively, which is a significant improvement over all other methods. However, without KD, our model shows worse performance than NSLP-G in all criteria. These results clearly demonstrate that KD has a significant impact on our method.

\subsubsection{Frame Length Assessment.}
The outputs of each model are further evaluated over a range of frame lengths (or duration) to verify for error propagation~\cite{li2021multitask} and false decoding initiation described in Section~\ref{sec:intro}. Specifically, FGD of the generated sign pose sequences is measured and averaged over each frame length interval. In terms of error propagation, FGD of PT starts at 0.24 and eventually increases to 0.37, as shown in Table~\ref{tab:2}. The data augmentation options (FN\&GN) alleviate this, but it still exists. This demonstrates that error propagation exists in AR-based models as FGD keeps increasing with a longer frame length. On the other hand, NAR-SLP models resolve this issue and perform better at longer frame lengths. 

We also measure FGD of each model at an early stage of prediction to verify false decoding initiation in NAR-based models. As shown in Table~\ref{tab:2}, FGD of NSLP-G is much larger than that of other models, showing the negative effect of the NAR method to some extent. However, our method alleviates the problem dramatically by reducing it from 0.50 to 0.09. This fact verifies that KD effectively resolves the false decoding initiation. Furthermore, both PT model outperforms the NAR-based model by achieving 0.04 and 0.05, respectively, because the AR-based models take the first sign pose and timing information from their ground truth during decoding~\cite{huang2021towards}. 

\subsection{Qualitative Results}
\begin{figure}[t!]
    \centering
    \includegraphics[width=1.0\textwidth]{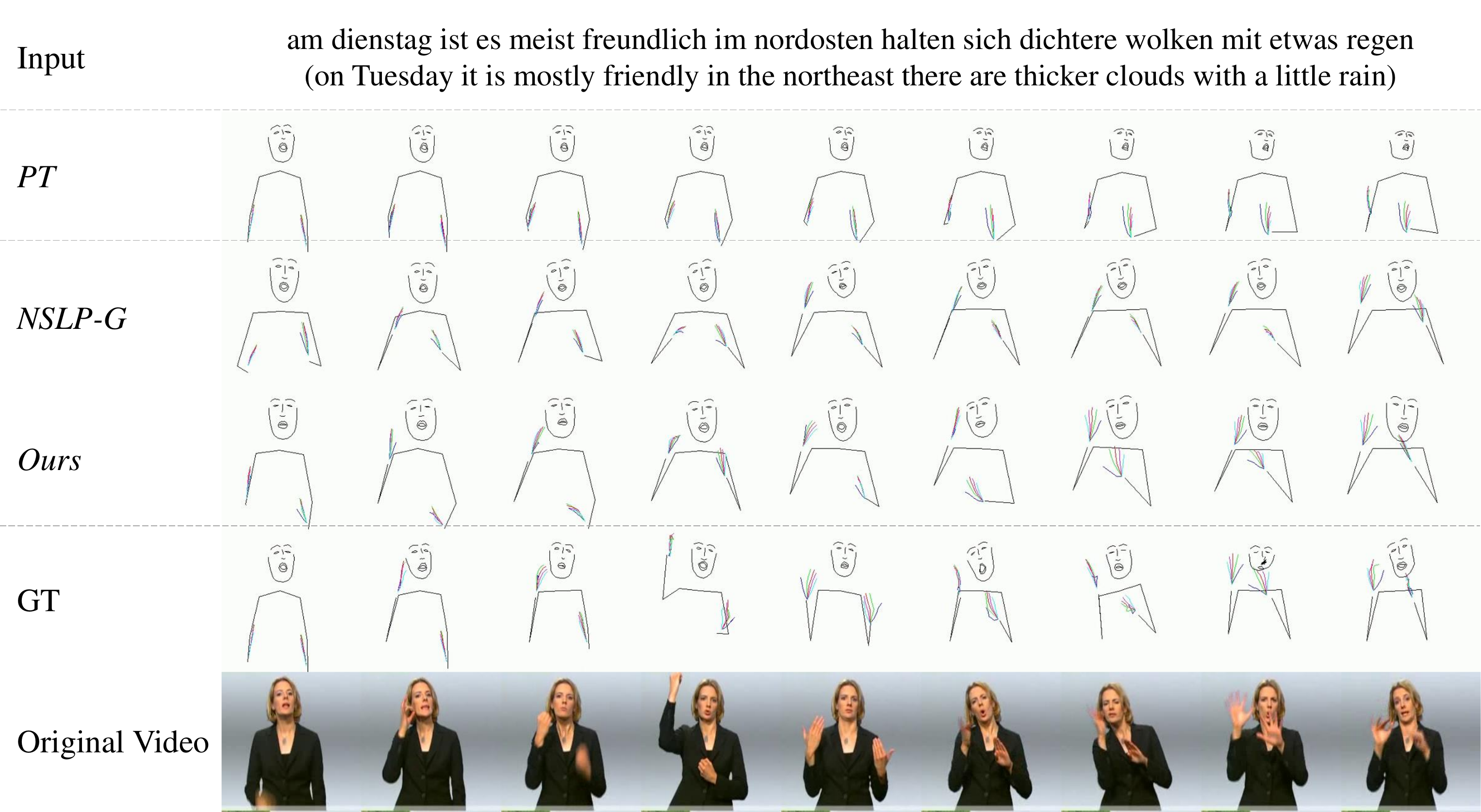}
    \caption{Visual comparison of generated sign poses over all methods.}
    \label{fig:fig2}
\end{figure}

The qualitative results of our model against other state-of-the-art models are shown in Figure~\ref{fig:fig2}. The first frame shows that PT starts with a given ground-truth frame but shows less dynamic and detailed outputs than other models. In contrast, NSLP-G is better at capturing frame-level movements without the error propagation problem. However, it starts with a poor sign pose and lacks detailed hand expressions within the overall sequence. Our approach shows a better start sign pose and more detailed hand expressions than NSLP-G. 

\subsection{Ablation Study} \label{sec:ablation}
To better understand our model in detail, we performed an ablation study to access several components, including KD and PLM, of our model in a controlled setting. 

\subsubsection{Loss study.}
We investigate the influence of the proposed loss formulation as in Equation~\ref{eq:loss}. We measure FGD and Duration Accuracy (DurACC) for the generated sign pose sequences and target frame length with the models trained with $\mathcal{L}_{pose}$ and $\mathcal{L}_{pose} + \mathcal{L}_{distil}$, respectively. Note that $\mathcal{L}_{length}$ is a mandatory loss of our model, as we design our model to generate both sign poses and length regulating ratio. DurACC assumes that the generated sign pose output is recorded at 25 FPS and measures the accuracy with less than a second margin of error. As shown in Table~\ref{subtab:loss_study}, a single loss is insufficient to optimize the model properly. On the other hand, using combined loss significantly improves the results in both FGD and DurACC. Thus, we conclude that $\mathcal{L}_{distil}$ plays a significant role in the multi-modal prediction of our model.

\subsubsection{Architecture design.}
We further experiment with different architectural choices to determine the impact of using PLM for multi-modal prediction. First, we implement a simple transformer as a baseline. Specifically, to implement the baseline, we use the [BOS] token for a pooling purpose to predict the target frame lengths of the given spoken language. Moreover, we use the same number of layers and dimensions as the BERT-based model for a fair comparison. As shown in Table~\ref{subtab:arch_study}, the performance results indicate that the model without BERT is far behind the model with BERT. Therefore, we conclude that the simple Transformer insufficient enough to learn two different modalities from the given input.

\begin{table}[t!]
    \begin{subtable}{.549\linewidth}
        \centering
        \setlength{\tabcolsep}{7pt}
        \resizebox{0.98\columnwidth}{!}{%
        \begin{tabular}{l|cc|cc}
        \toprule
        \multicolumn{1}{l|}{\multirow{3}{*}{Models}} & \multicolumn{2}{c|}{DEV}              & \multicolumn{2}{c}{TEST}              \\
        \multicolumn{1}{c|}{}                      & FGD$\dagger$           & \makecell{DurACC\\(\textless{}1s)} & FGD$\dagger$           & \makecell{DurACC\\(\textless{}1s)} \\ \midrule
        GT & 0.0  & 1.0  & 0.0  & 1.0  \\ \midrule
        $\mathcal{L}_{pose}+\mathcal{L}_{length}$ & 0.32 & 0.26 & 0.36 & 0.23 \\
        $\mathcal{L}_{pose}+\mathcal{L}_{length}+\mathcal{L}_{distil}$                                    & \textbf{0.15} & \textbf{0.76}         & \textbf{0.16} & \textbf{0.76}         \\ \bottomrule
        \end{tabular}
    }
    \subcaption{Effect of KD}
    \label{subtab:loss_study}
    \end{subtable}%
    \begin{subtable}{.451\linewidth}
        \centering
        \setlength{\tabcolsep}{7pt}
        \resizebox{0.98\columnwidth}{!}{%
        \begin{tabular}{l|cc|cc}
        \toprule
        \multicolumn{1}{l|}{\multirow{3}{*}{Models}} & \multicolumn{2}{c|}{DEV}              & \multicolumn{2}{c}{TEST}              \\
        \multicolumn{1}{c|}{}                        & FGD$\dagger$           & \makecell{DurACC\\(\textless{}1s)} & FGD$\dagger$           & \makecell{DurACC\\(\textless{}1s)} \\ \midrule
        GT       & 0.0 & 1.0 & 0.0 & 1.0 \\ \midrule
        w/o BERT & 0.27 & 0.50 & 0.30 & 0.52 \\
        w/ BERT                                      & \textbf{0.15} & \textbf{0.76}         & \textbf{0.16} & \textbf{0.76}         \\ \bottomrule
        \end{tabular}
    }
    \subcaption{Effect of PLM}
    \label{subtab:arch_study}
    \end{subtable} 
    \caption{Performance comparison between different loss types and architecture choices. Note that $\mathcal{L}_{length}$ is a mandatory loss of our model to regulate output length.}
    \label{tab:ablation_study}
\end{table}

\section{Discussion}

\subsection{Error Propagation Problem}
\begin{figure}[t!]
    \centering
    \includegraphics[width=0.9\columnwidth]{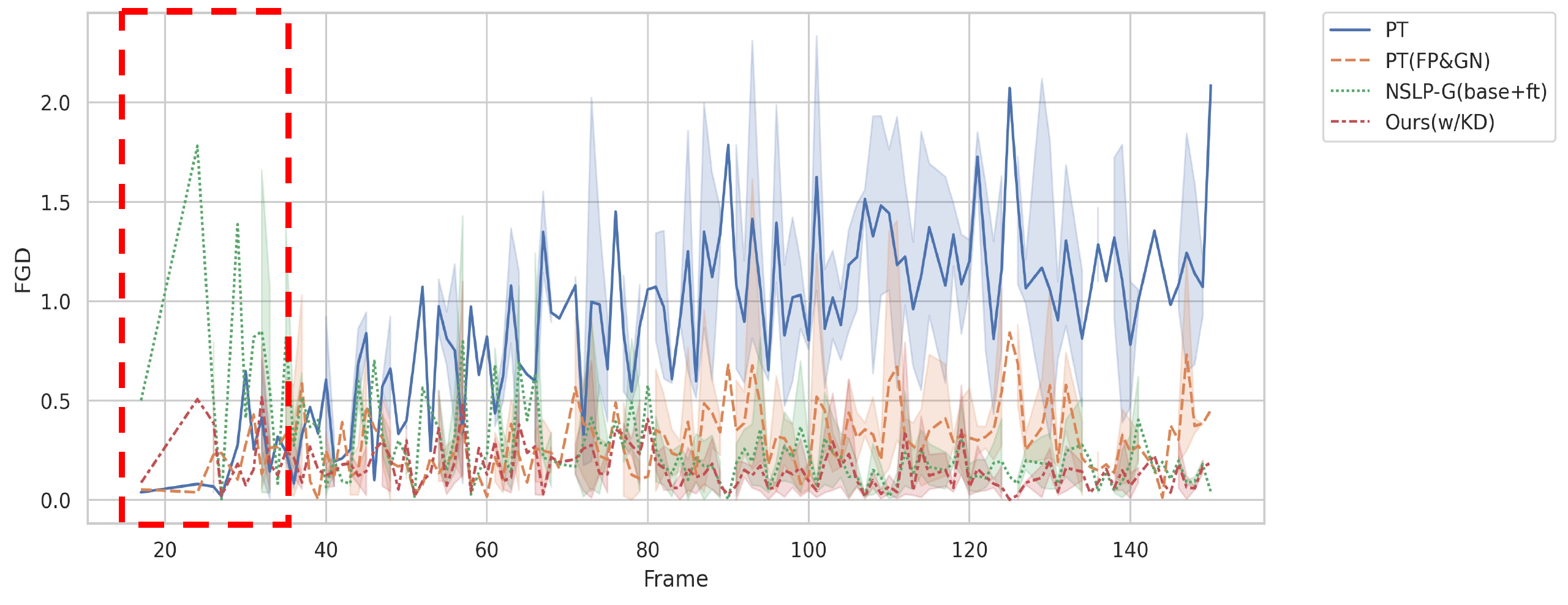}
    \caption{FGD line graph for each model according to frame length. The red dotted box indicates false decoding initiation in NAR models.}
    \label{fig:lineFGD}
\end{figure}

According to~\cite{li2021multitask}, AR-based models such as PT can easily propagate decoding errors to the next prediction. As shown in Figure~\ref{fig:lineFGD}, the FGD scores of PT models keep increasing with a longer frame length because the decoding errors continue to propagate. The data augmentation options (FN\&GN) alleviate this problem, but the problem still exists. This is because, to the best of our knowledge, AR-based models use a greedy decoding strategy to obtain sign pose sequences. On the other hand, NAR-based models such as NSLP-G and ours show consistent scores at longer frame lengths as the NAR decoder outputs sign pose sequence in parallel.

\subsection{Effect of Knowledge Distillation}
We further verify the false decoding initiation in the previous NAR model, which performs worse at shorter frame lengths. As shown in Figure~\ref{fig:lineFGD}, NSLP-G shows the worst performance in the early stage of prediction (red dotted box), which leads to overall performance degradation. In NSLP-G and our model, the pre-trained VAE is responsible for the spatial-linguistic features of SL and flows this information to the NAR Transformer. The previous model uses only a pre-trained decoder to align $z_{lang}$ indirectly, resulting in false decoding initiation. On the other hand, the proposed model alleviates this problem by a large margin by introducing KD that aligns directly with $z_{pose}$. Therefore, we conclude that KD is effective in NAR-based SLP models.

\subsection{Limitation of Back-Translation Evaluation}
Surprisingly, as shown in Table~\ref{tab:tab1}, our best performance surpasses the GT performance in the BT evaluation setting. It demonstrates that BT is limited in measuring the performance of the SLP models. BT also does not give consistent results when comparing PT with Mean and Random models as PT performs better in BLEU but worse in FGD. This is because BT strongly depends on the performance of the SLT model, and the translation performance is not yet stable enough to warrant the SLP models~\cite{Hwang2021nonautoregressive}. The evaluation results may become more accurate with the recent SLT model such as STMC~\cite{yin2020better}, suggesting that BT cannot be an absolute metric for SLP. However, at least BT serves as an indicator of how well the sign poses are generated in the evaluation results.

\section{Conclusions}
In this work, we proposed a novel Non-AutoRegressive Sign Language Production (NAR-SLP) via Knowledge Distillation (KD) to address the limitations of the existing SLP models, i.e., absence of target length prediction, error propagation, false decoding initiation, and less detailed outputs. We design our model to predict a sign pose sequence with its length. Furthermore, KD is adopted to directly align the outputs of a NAR Transformer with the pre-trained pose encoder. As a result, it alleviates the problem of false decoding initiation as well as the performance degradation due to the increased modality in our model. Moreover, our model introduces a multivariate Bernoulli distribution for sign poses, achieving a more detailed output. Extensive experiments have demonstrated the superior performance of the proposed approach. One of the particularly attractive properties of our method is that it learns the spatial and temporal aspects of Sign Language (SL) separately. Future work may therefore exploit our model to further improve the spatial aspect by introducing a graph representation~\cite{yan2018spatial}. We also plan to apply our model to a new benchmark SL dataset~\cite{duarte2021how2sign}. The code will be made publicly available.

%
%
\bibliographystyle{splncs04}
\bibliography{egbib}
\end{document}